\documentclass[letterpaper,10pt,conference]{ieeeconf}

\usepackage[table,xcdraw,dvipsnames]{xcolor}

\usepackage{cite}
\usepackage{amsmath,amssymb,amsfonts}
\usepackage{algorithmic}
\usepackage{graphicx}
\usepackage{textcomp}
\usepackage{subfiles}
\usepackage{svg}

\usepackage{times}
\usepackage{epsfig}
\usepackage{graphicx}
\usepackage{caption}
\usepackage{amsmath}
\usepackage{amssymb}
\usepackage{array,multirow,booktabs}
\usepackage{capt-of,etoolbox}
\usepackage{amsfonts}

\usepackage{cite}
\usepackage{amsmath,amssymb,amsfonts}
\usepackage{algorithmic}
\usepackage{graphicx}
\usepackage{textcomp}

\usepackage[breaklinks,colorlinks]{hyperref}

\definecolor{somegray}{rgb}{0.5, 0.5, 0.5}
\newcommand{\darkgrayed}[1]{\textcolor{somegray}{#1}}
\makeatletter
\newcommand*\titleheader[1]{\gdef\@titleheader{#1}}
\AtBeginDocument{%
  \let\st@red@title\@title
  \def\@title{%
    \vskip-3em
    \bgroup\normalfont\large\centering\@titleheader\par\egroup
    \vskip1.5em\st@red@title}
}
\makeatother

\titleheader{\darkgrayed{This paper has been accepted for publication at the \\
IEEE International Conference on Robotics and Automation (ICRA)), Atlanta, 2025.
\copyright IEEE}}

\title{\textbf{HS-SLAM: Hybrid Representation with Structural Supervision} \\ \textbf{for Improved Dense SLAM}
}

\author{Ziren Gong$^1$ \hspace{0.5cm} Fabio Tosi$^1$ \hspace{0.5cm} Youmin Zhang$^2$ \hspace{0.5cm} Stefano Mattoccia$^1$ \hspace{0.5cm} Matteo Poggi$^1$ \\
$^1$University of Bologna \hspace{0.5cm} $^2$Rock Universe %
\\
Project page: \url{https://zorangong.github.io/HS-SLAM/}}

\begin{document}

\twocolumn[{
\renewcommand\twocolumn[1][]{#1}
\maketitle
\begin{center}
    \vspace{-0.8cm}
    \begin{tabular}{l c c c}
        \includegraphics[width=\linewidth,scale=1.00]{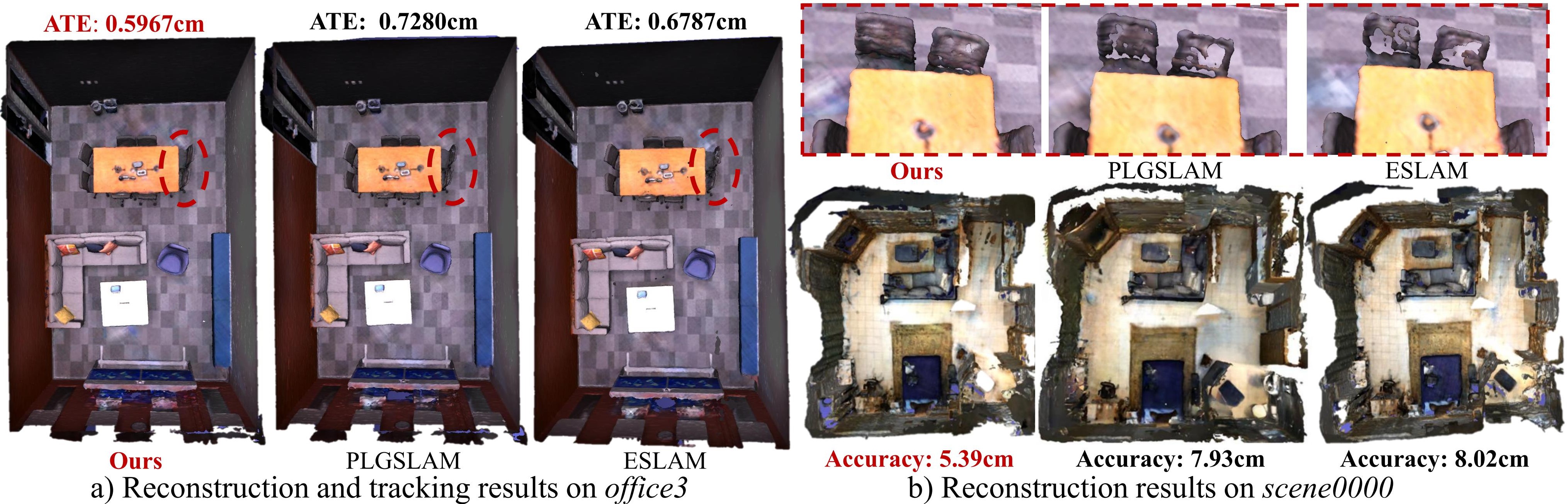}
    \end{tabular}
    \label{fig:teaser}
\end{center}
\small \hypertarget{fig:teaser}{Fig. 1.} \textbf{Trajectory errors on \emph{Office3} (Replica \cite{straub2019replica}) and 3D Reconstruction results on \emph{scene0000} (ScanNet \cite{dai2017scannet})}. Our framework yields the accurate 3D reconstruction shown in a), together with more precise camera tracking. Compared to existing methods such as PLGSLAM and ESLAM, HS-SLAM better preserves details (see \textcolor{red}{red circles and boxes}), achieving superior reconstruction accuracy, as shown in b).

\vspace{0.2cm}
}]

\begin{abstract}
NeRF-based SLAM has recently achieved promising results in tracking and reconstruction. However, existing methods face challenges in providing sufficient scene representation, capturing structural information, and maintaining global consistency in scenes emerging significant movement or being forgotten.
To this end, we present HS-SLAM to tackle these problems. To enhance scene representation capacity, we propose a hybrid encoding network that combines the complementary strengths of hash-grid, tri-planes, and one-blob, improving the completeness and smoothness of reconstruction. Additionally, we introduce structural supervision by sampling patches of non-local pixels rather than individual rays to better capture the scene structure. 
To ensure global consistency, we implement an active global bundle adjustment (BA) to eliminate camera drifts and mitigate accumulative errors. Experimental results demonstrate that HS-SLAM outperforms the baselines in tracking and reconstruction accuracy while maintaining the efficiency required for robotics.  
\end{abstract}

\section{Introduction}

Visual simultaneous localization and mapping (SLAM) has been widely applied in robotics. In the last decades, numerous traditional SLAM frameworks have achieved camera tracking and sparse maps, with hand-crafted methods \cite{newcombe2011kinectfusion, mur2017orb, li2021po, campos2021orb, li2024cto, newcombe2011dtam, salas2013slam++, whelan2015elasticfusion} demonstrating impressive scalability and fast speed. Deep-learning \cite{tateno2017cnn, li2020structure, teed2021droid} frameworks further enhance the accuracy and robustness of pose estimation. However, robots need dense reconstruction capacity with various applications of indoor reconstruction and virtual/augmented reality \cite{tang2020review}. With the emergence of NeRF \cite{mildenhall2021nerf} and 3D Gaussian Splatting (3DGS) \cite{kerbl20233d}, recent attention has turned to both NeRF-centric and 3DGS-centric SLAM.

3DGS-centric SLAM represents scenes with explicit 3D Gaussian-shaped primitives. SplaTAM \cite{keetha2024splatam} is the first open-source work to use the 3DGS \cite{kerbl20233d} in SLAM. GS-SLAM \cite{yan2024gs} also works out reasonably dense maps with 3D GS. Although these works demonstrate promising reconstruction capability, they face limitations concerning their huge memory usage and slow speed when handling both tracking and mapping simultaneously \cite{tosi2024nerfs}. These drawbacks hinder their applicability in robotics, where real-time performance and efficiency are essential.

Compared to 3DGS, NeRF-centric SLAM offers a more compact scene representation and requires significantly less GPU memory \cite{he2024nerfs}, making it highly suitable for resource-constrained robots. iMAP \cite{sucar2021imap} is the first work to represent scenes with implicit neural features. NICE-SLAM \cite{zhu2022nice}, ESLAM \cite{johari2023eslam}, and Co-SLAM \cite{wang2023co} further explore alternative representations to reconstruct high-fidelity scenes. Recently, the SOTA method PLGSLAM \cite{deng2024plgslam} further enhances the reconstruction accuracy in large-scale indoor scenes.

In this context, we investigate {whether NeRF-centric SLAM has achieved its optimal potential} from the following perspectives: 
i) scene representation, ii) structural supervision, and iii) global consistency. In terms of the former, the structure of embeddings plays a crucial role \cite{hua2024benchmarking}. Different encodings have distinct advantages: Tri-planes \cite{chan2022efficient} enable rapid convergence and enhance the completeness of unobserved regions, while hash grids \cite{muller2022instant} excel at capturing details.  
Regarding supervision, radiance-field-based approaches only learn from individual pixels, neglecting potential structural information in scenes, while for what concerns global consistency, the current NeRF-centric methods struggle with scenes being forgotten or 
in the occurrence of significant movements.

To this end, we propose HS-SLAM, a framework marrying a \underline{H}ybrid representation with \underline{S}tructural supervision for dense RGB-D SLAM. The former leverages the complementary advantages of hash grid \cite{muller2022instant}, tri-planes \cite{chan2022efficient} and one-blob \cite{muller2019neural}, achieving superior scene reconstruction. For the latter we incorporate patch rendering loss, enabling HS-SLAM to better capture structural features and perceive changes in unobserved regions. Finally, we introduce an active global BA to allocate more samples to historical observation where new regions emerge or scenes are being forgotten. This avoids re-training well-optimized scenes while eliminating camera drift and mitigating accumulative errors.

To resume, HS-SLAM envelops the following properties:

\begin{itemize}
\item \textbf{Enhanced scene representation capacity:} We introduce a novel hybrid representation to exploit complementary strengths of encodings, representing entire scenes with both completeness and detailed textures. 
\item \textbf{Non-local spatial and structural supervision:} We incorporate patch rendering loss for structural supervision, sampling patches of non-local pixels to capture structural information in scenes.  
\item \textbf{Global consistency:} Our framework proposes an active global BA strategy. It optimizes our SLAM system using historical observations while actively sampling from frames with new regions or scenes being forgotten.

\end{itemize}

\section{Related Work}

We briefly review the literature relevant to our work, referring to \cite{tosi2024nerfs3dgaussiansplatting} for a detailed overview of the latest advances.

\subsection{Traditional SLAM Frameworks}

These approaches generally refer to traditional visual SLAM pipelines, or to 
the deep-learning-based SLAM \cite{tateno2017cnn, li2020structure, teed2021droid} -- usually in a frame-to-frame fashion based on the visual features extracted from history frames. Furthermore, this kind of method commonly composes tracking, mapping, global BA, and loop closure. Traditional SLAM frameworks using depth points \cite{mur2017orb, campos2021orb, czarnowski2020deepfactors}, surfels \cite{schops2019bad}, and volumetric representations \cite{dai2017bundlefusion} achieve globally consistent reconstruction. However, their sparse and limited representation suffers from undesirable dense reconstruction results. 

\subsection{NeRF-centric Frameworks}

These approaches, also known as \textit{neural} SLAM systems, estimate camera poses and reconstruct dense maps by modeling scenes within MLPs and optimizing the scene representation. iMAP \cite{sucar2021imap} and Nice-SLAM \cite{zhu2022nice} are pioneers in bringing implicit neural mapping to SLAM. Point-SLAM \cite{sandstrom2023point} and Loopy-SLAM \cite{liso2024loopy} introduce a novel neural point embedding for dense reconstruction, allowing for the 
flexibility to correct and adjust local maps, yet with slow processing prohibiting deployment in robotics.  
ESLAM \cite{johari2023eslam} and Co-SLAM \cite{wang2023co} explore, respectively, tri-planes and hash grid embeddings for scene representation, improving both processing speed and reconstruction accuracy, with PLGSLAM \cite{deng2024plgslam} further improving the representation capacity in large indoor scenes. 
GO-SLAM \cite{zhang2023go} and KN-SLAM \cite{wu2024kn} exploit external trackers to achieve a good trade-off between camera poses and scene reconstruction.

Nevertheless, existing NeRF-centric methods still underutilize both the power of features representations \cite{hua2024benchmarking} and the structural supervision available from color images \cite{xie2023s3im}, thus attaining sub-optimal performance both at tracking and mapping.

\subsection{3DGS-centric Framworks}

These systems exploit explicit 3D Gaussian Splatting \cite{kerbl20233d} to represent the whole scene, 
iteratively growing the mapped area and offering the flexibility to adjust the reconstructed regions locally. SplaTAM \cite{keetha2024splatam} is the first open-source pioneer on this track,  
achieving high-quality color and depth rendering followed by GS-SLAM \cite{yan2024gs}. 
Although the 3DGS-centric frameworks prove the potential of 3DGS for SLAM applications, they are memory-hungry and run at a much lower speed, preventing their deployment in robotics.

\begin{figure*}[htbp]
	\centering
	\includegraphics[width=0.9\linewidth,scale=1.00]{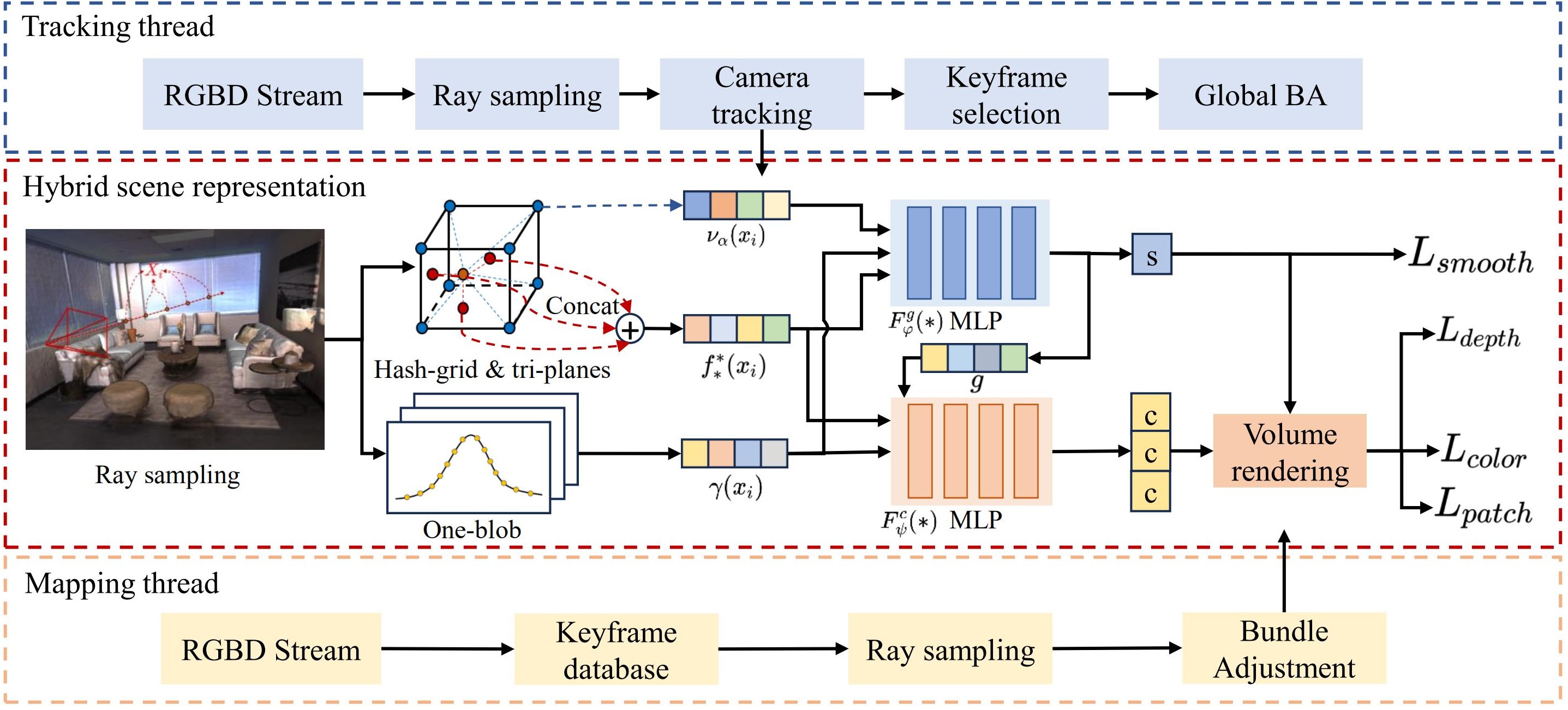}
	\caption{\textbf{The overview of HS-SLAM.} 1) Tracking thread: The proposed global BA actively optimizes our system taking ray sampling from historical observations. 2) Mapping thread: The system optimizes hybrid encodings by performing bundle adjustment. 3) Hybrid scene representation: We carefully designed the encoders composed of hash grid, tri-planes, and one-blob to improve completeness and accuracy. Meanwhile, the proposed structural supervision (patch rendering loss) allows the network to capture global scene structures.}
	\label{fig:hsslam}
\end{figure*}

\section{Method}

We introduce HS-SLAM, whose architecture is depicted in Fig. \ref{fig:hsslam}, from the following aspects: hybrid scene representation {\color{red} (III-A)}, SDF-based rendering and structural supervision {\color{red} (III-B)}, and active global bundle adjustment {\color{red} (III-C)}.

\subsection{Hybrid Scene Representation}\label{AA}
While \cite{johari2023eslam}, \cite{wang2023co}, and \cite{deng2024plgslam} further explore alternative representations, they rely solely on a single type of parameter encoding. 
However, each parameter encoding offers unique strengths: hash grids excel at capturing fine details while tri-planes effectively represent unobserved segments and enable rapid convergence \cite{hua2024benchmarking}. This insight naturally leads to the idea of combining their complementary advantages to achieve stronger scene representation. We propose a hybrid scene representation to improve the accuracy and completeness of reconstruction further. Hash grids and tri-planes are utilized as parametric encodings to capture high-frequency features \cite{wang2023co, johari2023eslam} like color and geometry details, while the one-blob technique is employed as a coordinate embedding to encode low-frequency information \cite{wang2023co}, such as coherence and smoothness priors.

Specifically, we employ a multi-resolution hash grid to encode various spatial features, capturing more detailed textures. Tri-linear interpolation is then applied to query the hash grid feature ${\large \nu_{\alpha } }\left ( x _{i} \right )$. For low-frequency features, we adopt the one-blob to encode spatial coordinates. The one-blob feature ${\large \gamma }\left ( x _{i} \right )$ encodes the smoothness and coherence priors to achieve high fidelity. To enhance stability in geometry reconstruction \cite{johari2023eslam}, two distinct tri-planes are designed at both coarse-level ${\large \tau ^{c}}\left ( x _{i}  \right )$ and fine-level ${\large \tau ^{f}  }\left ( x _{i}  \right )$, capturing appearance ${\large f _{\phi }^{a}}\left ( \ast  \right )$ and geometry features ${\large f _{\omega  }^{g}}\left ( \ast  \right )$ separately. The process of tri-plane encoding is formulated as follows:
\begin{gather}
    {\large f _{a}^{c}}\left ( x_{i} \right ) = \tau _{a-xy}^{c}\left ( x_{i} \right )  + \tau _{a-xz}^{c}\left ( x_{i} \right )+ \tau _{a-yz}^{c}\left ( x_{i} \right ) \notag \\
    {\large f _{a}^{f}}\left ( x_{i} \right ) = \tau _{a-xy}^{f}\left ( x_{i} \right )  + \tau _{a-xz}^{f}\left ( x_{i} \right )+ \tau _{a-yz}^{f}\left ( x_{i} \right ) \notag \\
    {\large f _{\phi }^{a}}\left ( x_{i} \right ) = Concat\left ( f_{a}^{c}, f_{a}^{f} \right ) \notag \\
    {\large f _{g}^{c}}\left ( x_{i} \right ) = \tau _{g-xy}^{c}\left ( x_{i} \right )  + \tau _{g-xz}^{c}\left ( x_{i} \right )+ \tau _{g-yz}^{c}\left ( x_{i} \right ) \notag \\
    {\large f _{g}^{f}}\left ( x_{i} \right ) = \tau _{g-xy}^{f}\left ( x_{i} \right )  + \tau _{g-xz}^{f}\left ( x_{i} \right )+ \tau _{g-yz}^{f}\left ( x_{i} \right ) \notag \\
    {\large f _{\omega  }^{g}}\left ( x_{i} \right ) = Concat\left ( f_{g}^{c}, f_{g}^{f} \right )
\end{gather}
where, ${\large f _{a}^{c}}$, ${\large f _{a}^{f}}$, ${\large f _{g}^{c}}$, ${\large f _{g}^{f}}$ represent the coarse and fine level of appearance features $f _{\phi }^{a}$ and geometry features $f _{\omega  }^{g}$ captured by tri-planes. The coordinates of the sampled points are denoted as $x_{i}$. $\left [ \tau _{a-xy}^{\ast}\left ( x_{i} \right ), \tau _{a-xz}^{\ast}\left ( x_{i} \right ), \tau _{a-yz}^{\ast}\left ( x_{i} \right )\right ]$ and $\left [ \tau _{g-xy}^{\ast}\left ( x_{i} \right ), \tau _{g-xz}^{\ast}\left ( x_{i} \right ), \tau _{g-yz}^{\ast}\left ( x_{i} \right )\right ]$ correspond to the three feature planes at both coarse and fine levels for appearance and geometry features, respectively. The two are then concatenated to form the final appearance and geometry features.

To learn the geometry features, our network combines the hash grid features ${\large \nu_{\alpha } }\left ( x _{i} \right )$, tri-planes features ${\large f _{\omega  }^{g}}\left ( x_{i} \right )$, and one-blob features ${\large \gamma }\left ( x _{i} \right )$. A shallow two-layer MLP $F_{\varphi }^{g}$ serves as the geometry decoder, which processes the hybrid encoding to 
predict signed distance function (SDF) $\phi _{g}\left ( x_{i}  \right ) $ and a geometry embedding $g$:
\begin{equation}
    F_{\varphi }^{g} \left ( {\large \nu_{\alpha } }\left ( x _{i} \right ), {\large f _{\omega  }^{g}}\left ( x_{i} \right ), {\large \gamma }\left ( x _{i} \right ) \right ) \to \left ( g, \phi _{g}\left ( x_{i}  \right )  \right )
\end{equation}

Then, a color decoder $F_{\psi  }^{c}$ inputs the appearance feature $f _{\phi }^{a}$, latent embedding $g$, and one-blob features ${\large \gamma }\left ( x _{i} \right )$ to predict the raw color $\phi _{a}\left ( x_{i}  \right ) $, with $f _{\phi }^{a}$ easing convergence, $g$ capturing detailed texture information, and ${\large \gamma }\left ( x _{i} \right )$ enhancing the smoothness and coherence of the predicted results. 
\begin{equation}
    F_{\psi  }^{c}\left ( {\large f _{\phi }^{a}}\left ( x_{i} \right ), {\large \gamma }\left ( x _{i} \right ), g \right ) \to \phi _{a}\left ( x_{i}  \right )  
\end{equation}

\subsection{SDF-based Rendering and Structural Supervision}
\textbf{SDF-based Rendering}. Our framework emulates the ray-casting process used in NeRF \cite{mildenhall2021nerf}. We implement different pixel sampling strategies for selecting rays depending on the specific component. In the global BA, detailed in (III-C), the network randomly samples pixels and their corresponding rays from historical observations. In the mapping thread, pixels are sampled from a window of frames, which includes the current frame, the adjacent frame, and additional frames randomly selected from keyframes. For all sampled points along the rays $\left \{ p_{n}  \right \}_{n=1}^{N}$, we query TSDF $\phi _{g}\left ( p_{n}  \right )$ and raw color $\phi _{a}\left ( p_{n}  \right )$ from our implicit neural network. SDF-based rendering \cite{or2022stylesdf} is then applied to calculate volume densities:
\begin{equation}
    \sigma \left ( p_{n}  \right ) = \beta \cdot Sigmoid \left (- \beta \cdot \phi _{g}\left ( p_{n}  \right ) \right )
\end{equation}
where $\sigma \left ( p_{n}  \right )$ denotes the volume density, and $\beta$ is a learnable parameter controlling the sharpness of the surface boundary. Then, the termination probability $w_{n}$, color $c$, and depth $d$ can be rendered by the calculated volume densities:
\begin{gather}
    w_{n} = exp\left ( -\sum_{k=1}^{n-1} \sigma \left ( p_{k}  \right )  \right )\left ( 1-exp\left ( -\sigma \left ( p_{n}  \right )  \right )  \right ) \notag \\
    \text{with} \quad\quad c  = \sum_{n=1}^{N} w_{n} \phi _{a}\left ( p_{n}  \right ), \quad\quad
    d  = \sum_{n=1}^{N} w_{n} z_{n} 
\end{gather}
where $z_{n}$ represents the depth of sampled points $p_{n}$.

\begin{figure}[t]
	\centering
	\includegraphics[width=\columnwidth,scale=1.00]{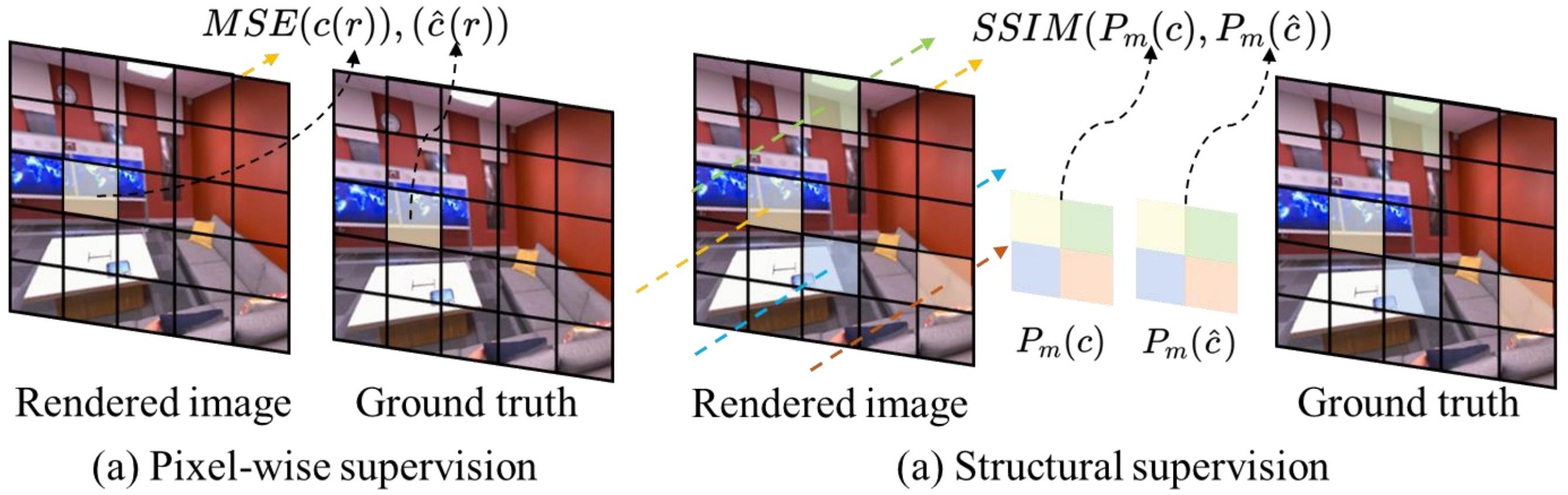}
 	\caption{\textbf{Differences between the pixel-wise and structural supervision. }This figure illustrates the key process of the pixel-wise and structural supervision. Our structural supervision captures non-local information from scenes. }
	\label{fig:s3im}
\end{figure}

\textbf{Patch-Based Structural Supervision}. Existing methods \cite{johari2023eslam, wang2023co, deng2024plgslam} only sample individual pixels and their corresponding rays separately. 
As illustrated in Fig. \ref{fig:s3im}, pixel-wise supervision employs MSE losses among individual pixels. 
However, this approach neglects any structural information. In contrast, we revise this paradigm 
to leverage the Structural Similarity Index Measure (SSIM) to capture structural information \cite{xie2023s3im}. 
Specifically, our network randomly selects $r$ pixels from the rendered set $R$ and their corresponding ground truth $R$ in the current frame or the keyframe database. The selected $r$ rendered pixels and their ground truth then form patches $P\left ( c\left ( r \right )  \right )$ and  
$P\left ( \hat{c} \left ( r \right )  \right )$, with $c\left ( r \right )$, $\hat{c} \left ( r \right )$ denoting rendered and ground truth color.

Afterward, we use a $3\times 3$ kernel and a stride factor of $4$ to compute SSIM.
The patch sampling and SSIM estimation steps are repeated $M$ times. The patch rendering loss is computed by the average of $M$ estimated SSIM:
\begin{equation}
    L_{patch}= 1 - \frac{1}{M}\sum_{m=1}^{M}SSIM\left ( P_{m}\left ( \hat{c}  \right ), P_{m}\left ( c  \right ) \right )
\end{equation}
We also compute color and depth rendering loss $L_{color}, L_{depth}$ between rendered values $c_{n}, d_{n}$ and ground truth values $\hat{c}_{n}, \hat{d}_{n}$ with $L_{2}$ errors.
\begin{equation}
    L_{color}=\frac{1}{N}\sum_{n=1}^{N}\left ( \hat{c}_{n}  - c_{n} \right )^{2}, L_{depth}=\frac{1}{N}\sum_{n=1}^{N}\left ( \hat{d}_{n}  - d_{n} \right )^{2}        
\end{equation}

\textbf{Other Losses.} In addition, we also enforce a feature smoothness loss to decrease noisy representation generated by hash grid ${\large \nu_{\alpha } }\left ( x _{i} \right )$ features in unobserved regions \cite{wang2023co}:
\begin{equation}
    L_{smooth}=\frac{1}{\left | X \right | } \sum_{x\in X} \bigtriangleup_{x}^{2} +\bigtriangleup_{y}^{2}+\bigtriangleup_{z}^{2} 
\end{equation}
with $X$ denoting a small random region to perform this loss, $\bigtriangleup _{x,y,z} = \nu_{\alpha }  \left ( x+\varepsilon_{x,y,z}   \right ) - \nu _{\alpha }\left ( x \right )$ representing feature-metric differences between adjacent sampled vertices on the hash grid along the three dimensions. 

We approximate the signed distance field as the following objective function:
\begin{equation}
    L_{sdf} \left ( P_{r}^{T}  \right ) = \frac{1}{\left | R \right | } \sum_{r\in R} \frac{1}{\left | P_{r}^{T} \right | }\sum_{p\in P_{r}^{T}} \left ( \phi _{g}\left ( p  \right )  \cdot T+ z\left ( p \right ) -D\left ( r \right )  \right )
\end{equation}
with $P_{r}^{T}$ denoting a group of points along the ray that lies within the truncation region, i.e. $\left | z\left ( p \right ) - D\left ( r \right )  \right | <  T$, $z\left ( p \right )$ being the planar depth of point $p$, and $D\left ( r \right )$ representing the measured ray depth. 
Furthermore, we distinguish the importance of points closer to the surface, situated in the middle of the truncation region $P_{r}^{Tm}$, from those located at the tail of the truncation region $P_{r}^{Tt}$:
\begin{equation}
    L_{sdfm}=L_{sdf}\left ( P_{r}^{Tm} \right ),  L_{sdft}=L_{sdf}\left ( P_{r}^{Tt} \right )
\end{equation}
Finally, for points sampled far from the surface $\left | z\left ( p \right )  - D\left ( r \right )  \right | \ge T$,
we enforce the predicted SDF values to be the truncated distance:
\begin{equation}
    L_{fs} = \frac{1}{\left | R \right | } \sum _{r\in R} \frac{1}{\left | P_{r}^{fs}  \right | } \sum_{p\in P_{r}^{fs} } \left ( \phi _{g}\left ( p  \right ) -1 \right )^{2}  
\end{equation}

\begin{table*}[t]
\caption{\textbf{Reconstruction and tracking results on Replica (mesh culling strategy in \cite{wang2023co}).} We compare our HS-SLAM with NeRF-centric SLAM systems that better meet efficiency requirements. }
\label{tab:table1}
\centering
\resizebox{0.85\textwidth}{!}{%
\begin{tabular}{lcccc|cc}
\hline
              & Acc.(cm)      & Comp.(cm)     & Comp.Rate($\%$)      & Depth l1(cm)  & Mean(cm)   & ATE RMSE(cm)       \\ \hline
Co-SLAM & 2.10  & 2.08 & 93.44 & 1.51 & 0.935 & 1.059 \\
ESLAM   & 2.18 & 1.75 & 96.46 & 0.94 & 0.565 & 0.707 \\
PLGSLAM & 2.15 & 1.74 & 96.25 & 0.83 & 0.525 & 0.635 \\
\textbf{HS-SLAM (ours)} & \textbf{1.96} & \textbf{1.67} & \textbf{96.57} & \textbf{0.71} & \textbf{0.463} & \textbf{0.528} \\ 
\hline
SplaTAM & - & - & - & - & - & 0.36- \\ \hline
\end{tabular}%
}
\end{table*}

\begin{figure*}[t]
	\centering
	\includegraphics[width=0.9\linewidth,scale=1.00]{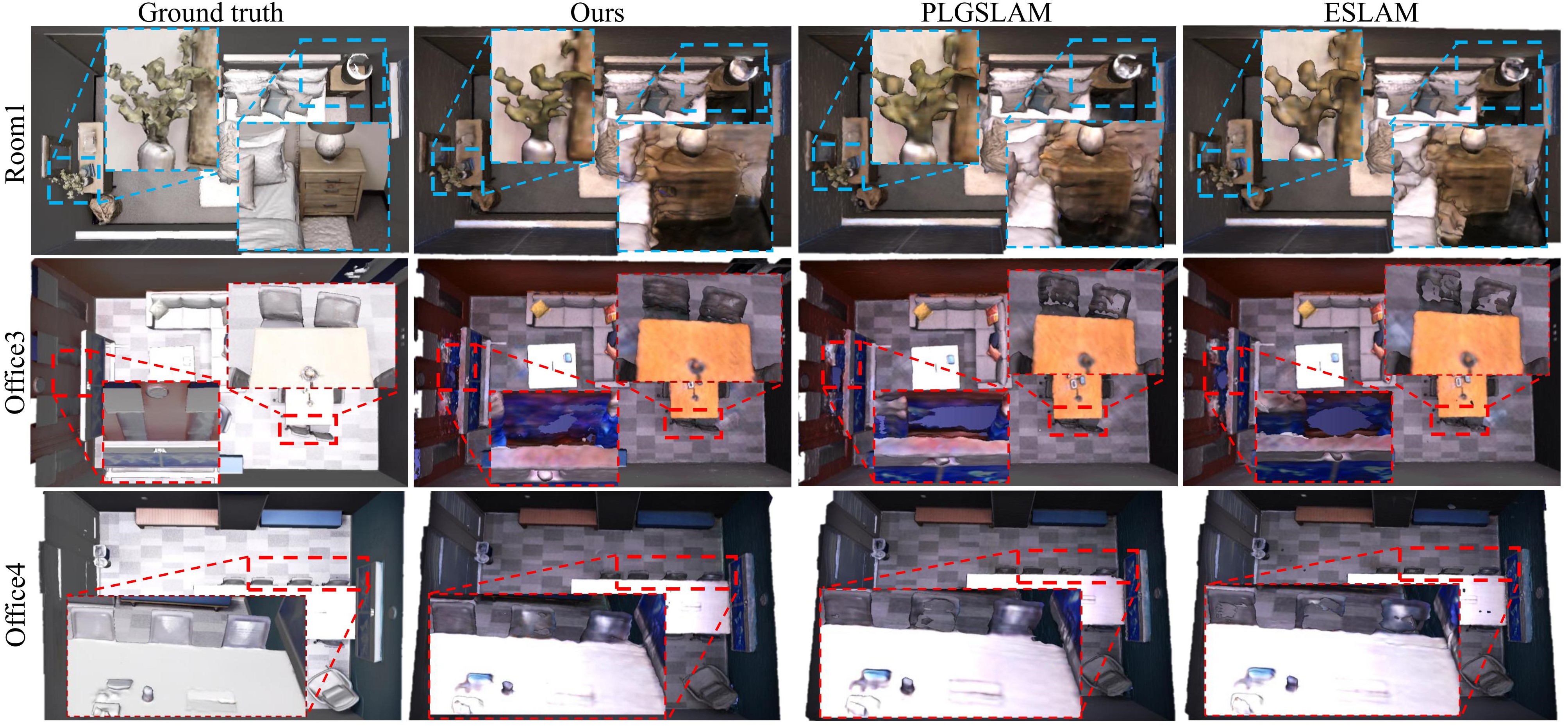}
 	\caption{\textbf{Qualitative reconstruction results on Replica. } For each scene, we provide detailed drawings, highlighting complex regions to emphasize the \textcolor{BlueGreen}{textures} and \textcolor{red}{completeness} of our reconstructed scenes. HS-SLAM demonstrates superior scene representation capabilities, effectively capturing intricate textures (e.g., \textcolor{BlueGreen}{flowers} and \textcolor{BlueGreen}{nightstands}) and achieving higher completeness in challenging areas (e.g., \textcolor{red}{chairs} and \textcolor{red}{walls}).}
	\label{fig:replica}
\end{figure*}

\subsection{Active Global Bundle Adjustment}

While existing NeRF-centric frameworks employ global BA to mitigate camera drifts, they select keyframes based on a fixed stride size, overlooking frames with significant motion and large trajectory errors. Meanwhile, processing well-optimized scenes again wastes lots of memory. To address this problem, we propose an active global BA to enhance the global consistency of both tracking and reconstruction. Our global BA actively selects the frames with suboptimal trajectory poses and scene reconstruction from history observations. We observe the fact that the global loss of frames increases markedly when the camera undergoes significant motion or the network starts to forget scenes. Consequently, a keyframe is selected when the global loss exceeds a threshold $L>t_{l}$ (we set $t_{l}=0.09$ ). Additionally, we aim to allocate more samples from the frames with higher losses.
To this end, we rank the keyframes by loss and sample rays from the top $N$ keyframes (with $N=15$). This approach prioritizes the keyframes inaccurately rendered by the model, which refers to newly explored frames or frames that the network is starting to forget. By randomly sampling rays from this prioritized global keyframe database, we optimize the global history trajectory, correcting inaccurate camera poses. 

\section{Experiments}

We now introduce our experimental evaluation, by detailing our setting and implementation details, then discussing the results achieved over several datasets.

\subsection{Experimental setup}\label{SCM}

We introduce the datasets used in our evaluation, as well as the frameworks we compare with, the evaluation metrics, and the implementation details of HS-SLAM.

\textbf{Datasets}. To evaluate the tracking and reconstruction accuracy, we test our HS-SLAM on several specific scenes in the following datasets: Replica \cite{straub2019replica}, ScanNet \cite{dai2017scannet}, and TUM RGB-D \cite{sturm2012benchmark}. The former is a semi-synthetic dataset allowing for the evaluation of both aspects involved in SLAM performance, whereas the others are usually employed to evaluate trajectory accuracy only.

\textbf{Baselines.} Considering the efficiency required in robotics applications, we focus our comparison with NeRF-centric SLAM systems achieving a reasonable framerate, such as Co-SLAM\cite{wang2023co}, ESLAM\cite{johari2023eslam}, and PLGSLAM \cite{deng2024plgslam} -- the current state-of-the-art NeRF-centric framework on Replica -- as the primary baselines to compare the accuracy of reconstruction and tracking with, while reporting SplaTAM as a reference for 3DGS-centric systems. When focusing on performance analysis, we compare our HS-SLAM also with 
Point-SLAM \cite{sandstrom2023point} to demonstrate our efficiency.

\textbf{Metrics.} To evaluate reconstruction quality, we use standard metrics such as \emph{depth l1} ($cm$), \emph{accuracy} ($cm$), \emph{completion} ($cm$), and \emph{completion rate} ($\%$). Unlike some prior works \cite{zhu2022nice, johari2023eslam, deng2024plgslam}, we adopt the mesh culling strategy proposed in \cite{wang2023co}, which minimizes artifacts outside the region of interest (ROI) and preserves occluded parts within it. This strategy includes frustum culling, occlusion handling, and the use of virtual cameras, covering occluded parts within the ROI. For camera tracking, we compute \emph{ATE RMSE} ($cm$).

\textbf{Implementation Details.} We implement HS-SLAM using PyTorch, on a desktop PC with NVIDIA RTX 3090 Ti GPU. We sample 2000 pixels for tracking and 4000 pixels for mapping. We use 16 bins for one-blob and a 16-level hash grid with a maximum voxel size of 2 cm. The tri-planes are set at 24cm/6cm and 24cm/3cm resolutions, each with 32 channels. The decoders consisted of two-layer MLPs with 32 channels in the hidden layers.

\begin{table*}[t]
\caption{\textbf{Ablation study on Replica.} We study the impact of any of our components on mapping and tracking accuracy.}
\label{tab:ablation}
\centering
\resizebox{0.9\textwidth}{!}{%
\begin{tabular}{clcccc|cc}
\hline
              && Acc.(cm)      & Comp.(cm)     & Comp.Rate($\%$) & Depth l1(cm)  & Mean(cm)       & ATE RMSE(cm)   \\ \hline
& \textbf{HS-SLAM (ours)} & \textbf{1.96} & \textbf{1.67} & 96.57  & \textbf{0.71} & \textbf{0.463} & \textbf{0.528} \\ \hline
(A) &Only hash grid($\times2$)      & 1.98     & 2.09     & 93.57    & 1.44  & 0.944  & 1.073 \\
(B) &Only tri-planes($\times2$)     & 2.05     & 1.68     & 96.68    & 0.81   & 0.476   & 0.547  \\
(C) & dense + tri-planes \cite{hua2024benchmarking}             & 5.61     & 1.73     & \textbf{97.03}    & 0.83   & 0.474   & 0.555  \\
\hline
(D) &w/o hash grid       & 2.06 & 1.68      & 96.55     & 0.75  & 0.487 & 0.558      \\
(E) &w/o tri-planes      & 2.11 & 1.93      & 94.52     & 1.21  & 0.851 & 0.915      \\
(F) &w/o one-blob        & 1.98 & 1.68      & 96.53     & 0.74  & 0.472 & 0.539      \\
\hline
(G) &w/o patch loss & 1.97 & 1.67      & 96.51     & 0.73  & 0.476 & 0.547 \\
(H) &w/o global BA       & 2.07 & 1.67      & 96.56     & 0.73  & 0.495 & 0.568 \\
(I) &w/ Co-SLAM global BA & 2.01 & 1.67      & 96.57     & 0.73  & 0.485 & 0.557\\
\hline
\end{tabular}%
}
\end{table*}

\subsection{Results on Replica}
We evaluate our HS-SLAM on Replica, using the same RGB-D sequences as the baselines to test reconstruction accuracy. 
Tab. \ref{tab:table1} shows how HS-SLAM achieves both superior reconstruction accuracy, outperforming any NeRF-centric SLAM system using single parametric encoding, either hash grids (Co-SLAM) or tri-planes (ESLAM, PLGSLAM) thanks to the hybrid encoding. Furthermore, our global BA strategy based on active sampling allows for consistently superior tracking accuracy.
Fig. \ref{fig:replica} shows a qualitative comparison between HS-SLAM, PLGSLAM, and ESLAM, highlighting the lower presence of artifacts in our reconstructions.

\subsection{Ablation Studies}
We assess the impact of any components in HS-SLAM on the final accuracy, collecting the results in Tab. \ref{tab:ablation}.

\begin{table}[t]
\caption{\textbf{Tracking and mapping results on ScanNet.}
}
\label{tab:scannet}
\resizebox{\columnwidth}{!}{%
\begin{tabular}{lccccccc}
\hline
              & Acc.(cm)       & Comp.(cm)     & Comp.Rate($\%$) & ATE RMSE(cm)  \\ \hline
Co-SLAM       & 31.31          & 3.83          & \textbf{77.36}  & 9.37          \\
ESLAM         & 25.96          & 4.27          & 73.58           & 7.52          \\
PLGSLAM       & 21.14          & 4.29          & 74.78           & \textbf{6.77} \\
\textbf{HS-SLAM (ours)} & \textbf{19.34} & \textbf{4.25} & 73.33           & 7.42          \\ \hline
SplaTAM & - & - & - & 11.88 \\ \hline
\end{tabular}%
}
\end{table}

\begin{table}[t]
\caption{\textbf{Tracking results on TUM RGB-D.} }
\label{tab:tum}
\resizebox{\columnwidth}{!}{%
\begin{tabular}{lcccc}
\hline
              & fr1/desk(cm)   & fr2/xyz(cm)   & fr3/office(cm) & Average(cm)    \\ \hline
Co-SLAM       & 2.754          & 1.747         & 3.293          & 2.598          \\
ESLAM         & 2.519          & 1.212         & 2.786          & 2.172          \\
\textbf{HS-SLAM (ours)} & \textbf{2.485} & \textbf{1.11} & \textbf{2.77}  & \textbf{2.122} \\ \hline
SplaTAM & 3.35- & 1.24- & 5.16- & 3.25- \\ \hline
\end{tabular}%
}
\end{table}

\textbf{Hybrid Scene Representation}. 
We compare several embedding variants aimed at finding the one best suited for SLAM. To stress the importance of combining different representations to complement their strengths and weaknesses, at the top of the table we report results achieved by a single embedding, hash grid (A), or tri-planes (B), by doubling the number of features they store. Both fail at obtaining satisfying results across the overall set of metrics, confirming that a hybrid encoding is crucial to overcome the limitations of the single ones -- and does not simply benefit from having more features.
We then evaluate the results achieved using the hybrid encoding proposed in \cite{hua2024benchmarking}, claiming that dense grids and tri-planes allow for the best results (C). Although yielding the highest completion rate, this solution performs worse on most of the remaining metrics, confirming our synergy between hash grid, tri-planes, and one-blob encoding as the optimal choice for hybrid encoding. This is further confirmed by removing a single among these representations (D-F), which always reduces the results on any metric.

\textbf{Structural Supervision}. The absence of structural supervision (G) introduces a moderate drop in mapping accuracy, as well as in overall tracking quality. This proves how, despite not playing a crucial role as hybrid embeddings, the structural supervision coming from patches is necessary to obtain optimal results.

\textbf{Active Global Bundle Adjustment}.
Removing global BA (H) yields drops both in mapping and tracking quality, highlighting its major role in reducing cumulative camera pose errors, specifically when handling newly explored scenes or scenes being forgotten. Finally, replacing it with Co-SLAM global BA (I) leads to poorer performance, demonstrating once more the advance introduced by our global BA.

\subsection{Results on Other Datasets}

\textbf{ScanNet Dataset}. We evaluate tracking and reconstruction accuracy on ScanNet, collecting the results in Tab. \ref{tab:scannet}. On the one hand, HS-SLAM outperforms other methods in reconstruction. On the other hand, PLGSLAM achieves better tracking results due to its design specifically tailored for larger scene reconstruction, a contribution orthogonal to ours. Nonetheless, HS-SLAM still demonstrates strong performance thanks to its accurate scene representation and high reconstruction completeness.  

\textbf{TUM Dataset}. We further test HS-SLAM on the TUM dataset and evaluate tracking accuracy. Tab. \ref{tab:tum} demonstrates that our framework achieves promising tracking accuracy, outperforming once again both Co-SLAM and ESLAM.

\begin{table}[t]
\caption{\textbf{Memory usage and runtime on Replica \emph{room0}}. Mem. Size sums the footprints by both encoder and decoder. Avg. FPS divides the total runtime by the number of frames.}
\label{tab:efficiency}
\resizebox{\columnwidth}{!}{%
\begin{tabular}{lccc}
\hline
           & Scene Encoding       & Mem.Size (MB) & Avg.FPS       \\ \hline
Co-SLAM    & Hash Grid + MLP      & \textbf{6.72} & \textbf{7.97} \\
ESLAM      & Feature Planes + MLP & 27.21         & 4.62          \\
PLGSLAM    & Feature Planes + MLP & -             & -             \\ 
\textbf{HS-SLAM} (ours)   & Hybrid + MLP    & 25.83         & 2.35          \\ \hline
Point-SLAM & Neural Points + MLP  & 54.8/3936.8         & 0.23          \\
SplaTAM    & 3D Gaussians         & 392.5         & 0.14          \\ \hline
\end{tabular}%
}
\end{table}

\subsection{Performance Analysis}
We compare our HS-SLAM with the NeRF-centric SLAM systems considered so far, as well as with Point-SLAM and SplaTAM, to assess its efficiency. For a fair comparison, we evaluate the available open-source code on our desktop PC. We measured memory usage (total footprint of the encoder and decoder) and average FPS (average time required to process per frame in a sequence). Tab. \ref{tab:efficiency} shows that Co-SLAM is the fastest method with the lowest model size, yet the least accurate according to the previous experiments. In contrast, Point-SLAM and SplaTAM demonstrate much lower speeds (less than 1 FPS) with a large model (hundreds or thousands of MB), making neural points and 3D Gaussian Splatting still unsuitable for real-time applications. 
HS-SLAM represents a good trade-off between the two worlds, outperforming Co-SLAM and ESLAM while retaining a few FPS in speed.

\section{Conclusion}
We present HS-SLAM, a dense RGB-D SLAM system that integrates hybrid scene representation with structural supervision, aided by an active global BA to mitigate camera drifts. Experimental results demonstrate HS-SLAM's superior performance in both reconstruction and pose estimation compared to existing NeRF-centric SLAM systems, achieving the best trade-off between accuracy and efficiency. While currently optimized for indoor environments, future integration of submaps and loop closure techniques could expand its capability to handle larger-scale scenes.

\footnotesize \textbf{Acknowledgment}. We sincerely thank the scholarship supported by China Scholarship Council (CSC).

\bibliographystyle{ieee_fullname}
\bibliography{references}

\begin{thebibliography}{10}\itemsep=-1pt

\bibitem{campos2021orb}
Carlos Campos, Richard Elvira, Juan J~G{\'o}mez Rodr{\'\i}guez, Jos{\'e}~MM
  Montiel, and Juan~D Tard{\'o}s.
\newblock Orb-slam3: An accurate open-source library for visual,
  visual--inertial, and multimap slam.
\newblock {\em IEEE Transactions on Robotics}, 37(6):1874--1890, 2021.

\bibitem{chan2022efficient}
Eric~R Chan, Connor~Z Lin, Matthew~A Chan, Koki Nagano, Boxiao Pan, Shalini
  De~Mello, Orazio Gallo, Leonidas~J Guibas, Jonathan Tremblay, Sameh Khamis,
  et~al.
\newblock Efficient geometry-aware 3d generative adversarial networks.
\newblock In {\em Proceedings of the IEEE/CVF conference on computer vision and
  pattern recognition}, pages 16123--16133, 2022.

\bibitem{czarnowski2020deepfactors}
Jan Czarnowski, Tristan Laidlow, Ronald Clark, and Andrew~J Davison.
\newblock Deepfactors: Real-time probabilistic dense monocular slam.
\newblock {\em IEEE Robotics and Automation Letters}, 5(2):721--728, 2020.

\bibitem{dai2017scannet}
Angela Dai, Angel~X Chang, Manolis Savva, Maciej Halber, Thomas Funkhouser, and
  Matthias Nie{\ss}ner.
\newblock Scannet: Richly-annotated 3d reconstructions of indoor scenes.
\newblock In {\em Proceedings of the IEEE conference on computer vision and
  pattern recognition}, pages 5828--5839, 2017.

\bibitem{dai2017bundlefusion}
Angela Dai, Matthias Nie{\ss}ner, Michael Zollh{\"o}fer, Shahram Izadi, and
  Christian Theobalt.
\newblock Bundlefusion: Real-time globally consistent 3d reconstruction using
  on-the-fly surface reintegration.
\newblock {\em ACM Transactions on Graphics (ToG)}, 36(4):1, 2017.

\bibitem{deng2024plgslam}
Tianchen Deng, Guole Shen, Tong Qin, Jianyu Wang, Wentao Zhao, Jingchuan Wang,
  Danwei Wang, and Weidong Chen.
\newblock Plgslam: Progressive neural scene represenation with local to global
  bundle adjustment.
\newblock In {\em Proceedings of the IEEE/CVF Conference on Computer Vision and
  Pattern Recognition}, pages 19657--19666, 2024.

\bibitem{he2024nerfs}
Siming He, Zach Osman, and Pratik Chaudhari.
\newblock From nerfs to gaussian splats, and back.
\newblock {\em arXiv preprint arXiv:2405.09717}, 2024.

\bibitem{hua2024benchmarking}
Tongyan Hua and Lin Wang.
\newblock Benchmarking implicit neural representation and geometric rendering
  in real-time rgb-d slam.
\newblock In {\em Proceedings of the IEEE/CVF Conference on Computer Vision and
  Pattern Recognition}, pages 21346--21356, 2024.

\bibitem{johari2023eslam}
Mohammad~Mahdi Johari, Camilla Carta, and Fran{\c{c}}ois Fleuret.
\newblock Eslam: Efficient dense slam system based on hybrid representation of
  signed distance fields.
\newblock In {\em Proceedings of the IEEE/CVF Conference on Computer Vision and
  Pattern Recognition}, pages 17408--17419, 2023.

\bibitem{keetha2024splatam}
Nikhil Keetha, Jay Karhade, Krishna~Murthy Jatavallabhula, Gengshan Yang,
  Sebastian Scherer, Deva Ramanan, and Jonathon Luiten.
\newblock Splatam: Splat track \& map 3d gaussians for dense rgb-d slam.
\newblock In {\em Proceedings of the IEEE/CVF Conference on Computer Vision and
  Pattern Recognition}, pages 21357--21366, 2024.

\bibitem{kerbl20233d}
Bernhard Kerbl, Georgios Kopanas, Thomas Leimk{\"u}hler, and George Drettakis.
\newblock 3d gaussian splatting for real-time radiance field rendering.
\newblock {\em ACM Trans. Graph.}, 42(4):139--1, 2023.

\bibitem{li2021po}
Xiaohan Li, Shiqi Lin, Meng Xu, Deyun Dai, and Jikai Wang.
\newblock Po-slam: A novel monocular visual slam with points and objects.
\newblock In {\em 2021 4th International conference on artificial intelligence
  and big data (ICAIBD)}, pages 454--458. IEEE, 2021.

\bibitem{li2024cto}
Xiaohan Li, Dong Liu, and Jun Wu.
\newblock Cto-slam: Contour tracking for object-level robust 4d slam.
\newblock In {\em Proceedings of the AAAI Conference on Artificial
  Intelligence}, volume~38, pages 10323--10331, 2024.

\bibitem{li2020structure}
Yanyan Li, Nikolas Brasch, Yida Wang, Nassir Navab, and Federico Tombari.
\newblock Structure-slam: Low-drift monocular slam in indoor environments.
\newblock {\em IEEE Robotics and Automation Letters}, 5(4):6583--6590, 2020.

\bibitem{liso2024loopy}
Lorenzo Liso, Erik Sandstr{\"o}m, Vladimir Yugay, Luc Van~Gool, and Martin~R
  Oswald.
\newblock Loopy-slam: Dense neural slam with loop closures.
\newblock In {\em Proceedings of the IEEE/CVF Conference on Computer Vision and
  Pattern Recognition}, pages 20363--20373, 2024.

\bibitem{mildenhall2021nerf}
Ben Mildenhall, Pratul~P Srinivasan, Matthew Tancik, Jonathan~T Barron, Ravi
  Ramamoorthi, and Ren Ng.
\newblock Nerf: Representing scenes as neural radiance fields for view
  synthesis.
\newblock {\em Communications of the ACM}, 65(1):99--106, 2021.

\bibitem{muller2022instant}
Thomas M{\"u}ller, Alex Evans, Christoph Schied, and Alexander Keller.
\newblock Instant neural graphics primitives with a multiresolution hash
  encoding.
\newblock {\em ACM transactions on graphics (TOG)}, 41(4):1--15, 2022.

\bibitem{muller2019neural}
Thomas M{\"u}ller, Brian McWilliams, Fabrice Rousselle, Markus Gross, and Jan
  Nov{\'a}k.
\newblock Neural importance sampling.
\newblock {\em ACM Transactions on Graphics (ToG)}, 38(5):1--19, 2019.

\bibitem{mur2017orb}
Raul Mur-Artal and Juan~D Tard{\'o}s.
\newblock Orb-slam2: An open-source slam system for monocular, stereo, and
  rgb-d cameras.
\newblock {\em IEEE transactions on robotics}, 33(5):1255--1262, 2017.

\bibitem{newcombe2011kinectfusion}
Richard~A Newcombe, Shahram Izadi, Otmar Hilliges, David Molyneaux, David Kim,
  Andrew~J Davison, Pushmeet Kohi, Jamie Shotton, Steve Hodges, and Andrew
  Fitzgibbon.
\newblock Kinectfusion: Real-time dense surface mapping and tracking.
\newblock In {\em 2011 10th IEEE international symposium on mixed and augmented
  reality}, pages 127--136. Ieee, 2011.

\bibitem{newcombe2011dtam}
Richard~A Newcombe, Steven~J Lovegrove, and Andrew~J Davison.
\newblock Dtam: Dense tracking and mapping in real-time.
\newblock In {\em 2011 international conference on computer vision}, pages
  2320--2327. IEEE, 2011.

\bibitem{or2022stylesdf}
Roy Or-El, Xuan Luo, Mengyi Shan, Eli Shechtman, Jeong~Joon Park, and Ira
  Kemelmacher-Shlizerman.
\newblock Stylesdf: High-resolution 3d-consistent image and geometry
  generation.
\newblock In {\em Proceedings of the IEEE/CVF Conference on Computer Vision and
  Pattern Recognition}, pages 13503--13513, 2022.

\bibitem{salas2013slam++}
Renato~F Salas-Moreno, Richard~A Newcombe, Hauke Strasdat, Paul~HJ Kelly, and
  Andrew~J Davison.
\newblock Slam++: Simultaneous localisation and mapping at the level of
  objects.
\newblock In {\em Proceedings of the IEEE conference on computer vision and
  pattern recognition}, pages 1352--1359, 2013.

\bibitem{sandstrom2023point}
Erik Sandstr{\"o}m, Yue Li, Luc Van~Gool, and Martin~R Oswald.
\newblock Point-slam: Dense neural point cloud-based slam.
\newblock In {\em Proceedings of the IEEE/CVF International Conference on
  Computer Vision}, pages 18433--18444, 2023.

\bibitem{schops2019bad}
Thomas Schops, Torsten Sattler, and Marc Pollefeys.
\newblock Bad slam: Bundle adjusted direct rgb-d slam.
\newblock In {\em Proceedings of the IEEE/CVF Conference on Computer Vision and
  Pattern Recognition}, pages 134--144, 2019.

\bibitem{straub2019replica}
Julian Straub, Thomas Whelan, Lingni Ma, Yufan Chen, Erik Wijmans, Simon Green,
  Jakob~J Engel, Raul Mur-Artal, Carl Ren, Shobhit Verma, et~al.
\newblock The replica dataset: A digital replica of indoor spaces.
\newblock {\em arXiv preprint arXiv:1906.05797}, 2019.

\bibitem{sturm2012benchmark}
J{\"u}rgen Sturm, Nikolas Engelhard, Felix Endres, Wolfram Burgard, and Daniel
  Cremers.
\newblock A benchmark for the evaluation of rgb-d slam systems.
\newblock In {\em 2012 IEEE/RSJ international conference on intelligent robots
  and systems}, pages 573--580. IEEE, 2012.

\bibitem{sucar2021imap}
Edgar Sucar, Shikun Liu, Joseph Ortiz, and Andrew~J Davison.
\newblock imap: Implicit mapping and positioning in real-time.
\newblock In {\em Proceedings of the IEEE/CVF international conference on
  computer vision}, pages 6229--6238, 2021.

\bibitem{tang2020review}
Baihui Tang and Sanxing Cao.
\newblock A review of vslam technology applied in augmented reality.
\newblock In {\em IOP Conference Series: Materials Science and Engineering},
  volume 782, page 042014. IOP Publishing, 2020.

\bibitem{tateno2017cnn}
Keisuke Tateno, Federico Tombari, Iro Laina, and Nassir Navab.
\newblock Cnn-slam: Real-time dense monocular slam with learned depth
  prediction.
\newblock In {\em Proceedings of the IEEE conference on computer vision and
  pattern recognition}, pages 6243--6252, 2017.

\bibitem{teed2021droid}
Zachary Teed and Jia Deng.
\newblock Droid-slam: Deep visual slam for monocular, stereo, and rgb-d
  cameras.
\newblock {\em Advances in neural information processing systems},
  34:16558--16569, 2021.

\bibitem{tosi2024nerfs}
Fabio Tosi, Youmin Zhang, Ziren Gong, Erik Sandstr{\"o}m, Stefano Mattoccia,
  Martin~R Oswald, and Matteo Poggi.
\newblock How nerfs and 3d gaussian splatting are reshaping slam: a survey.
\newblock {\em arXiv preprint arXiv:2402.13255}, 4, 2024.

\bibitem{tosi2024nerfs3dgaussiansplatting}
Fabio Tosi, Youmin Zhang, Ziren Gong, Erik Sandström, Stefano Mattoccia,
  Martin~R. Oswald, and Matteo Poggi.
\newblock How nerfs and 3d gaussian splatting are reshaping slam: a survey,
  2024.

\bibitem{wang2023co}
Hengyi Wang, Jingwen Wang, and Lourdes Agapito.
\newblock Co-slam: Joint coordinate and sparse parametric encodings for neural
  real-time slam.
\newblock In {\em Proceedings of the IEEE/CVF Conference on Computer Vision and
  Pattern Recognition}, pages 13293--13302, 2023.

\bibitem{whelan2015elasticfusion}
Thomas Whelan, Stefan Leutenegger, Renato~F Salas-Moreno, Ben Glocker, and
  Andrew~J Davison.
\newblock Elasticfusion: Dense slam without a pose graph.
\newblock In {\em Robotics: science and systems}, volume~11, page~3. Rome,
  Italy, 2015.

\bibitem{wu2024kn}
Xingming Wu, Zimeng Liu, Yuxin Tian, Zhong Liu, and Weihai Chen.
\newblock Kn-slam: Keypoints and neural implicit encoding slam.
\newblock {\em IEEE Transactions on Instrumentation and Measurement}, 73:1--12,
  2024.

\bibitem{xie2023s3im}
Zeke Xie, Xindi Yang, Yujie Yang, Qi Sun, Yixiang Jiang, Haoran Wang, Yunfeng
  Cai, and Mingming Sun.
\newblock S3im: Stochastic structural similarity and its unreasonable
  effectiveness for neural fields.
\newblock In {\em Proceedings of the IEEE/CVF International Conference on
  Computer Vision}, pages 18024--18034, 2023.

\bibitem{yan2024gs}
Chi Yan, Delin Qu, Dan Xu, Bin Zhao, Zhigang Wang, Dong Wang, and Xuelong Li.
\newblock Gs-slam: Dense visual slam with 3d gaussian splatting.
\newblock In {\em Proceedings of the IEEE/CVF Conference on Computer Vision and
  Pattern Recognition}, pages 19595--19604, 2024.

\bibitem{zhang2023go}
Youmin Zhang, Fabio Tosi, Stefano Mattoccia, and Matteo Poggi.
\newblock Go-slam: Global optimization for consistent 3d instant
  reconstruction.
\newblock In {\em Proceedings of the IEEE/CVF International Conference on
  Computer Vision}, pages 3727--3737, 2023.

\bibitem{zhu2022nice}
Zihan Zhu, Songyou Peng, Viktor Larsson, Weiwei Xu, Hujun Bao, Zhaopeng Cui,
  Martin~R Oswald, and Marc Pollefeys.
\newblock Nice-slam: Neural implicit scalable encoding for slam.
\newblock In {\em Proceedings of the IEEE/CVF conference on computer vision and
  pattern recognition}, pages 12786--12796, 2022.

\end{thebibliography}

\end{document}